\documentclass[10pt,twocolumn,letterpaper]{article}

\usepackage{cvpr}              

\usepackage{xspace}
\usepackage{multirow}
\usepackage{makecell}
\usepackage{amsmath,amssymb,bbm}
\definecolor{dgreen}{RGB}{0,150,0}

\newif\ifdraft
\draftfalse

\ifdraft
 \newcommand{\PF}[1]{{\color{red}{\bf PF: #1}}}
 \newcommand{\pf}[1]{{\color{red} #1}}
  \newcommand{\JC}[1]{{\color{blue}{\bf JC: #1}}} 
 
  \newcommand{\ME}[1]{{\color{dgreen}{\bf ME: #1}}}
 
\newcommand{\dm}[1]{{\color{violet} #1}}

 \newcommand{\TODO}[1]{\textbf{\color{yellow}[TODO: #1]}}
\else
 \newcommand{\PF}[1]{}
 \newcommand{\pf}[1]{#1}
 \newcommand{\JY}[1]{} 
 
 \newcommand{\WL}[1]{}
 
 \newcommand{\JC}[1]{}
 
 \newcommand{\ME}[1]{}

 \newcommand{\dm}[1]{}
 \newcommand{\TODO}[1]{}
 
\fi

\newcommand{\parag}[1]{\vspace{-3mm}\paragraph{#1}}

\newcommand\net{MedTet}

\definecolor{cvprblue}{rgb}{0.21,0.49,0.74}
\usepackage[pagebackref,breaklinks,colorlinks,allcolors=cvprblue]{hyperref}

\def\paperID{1226} 
\def\confName{CVPR}
\def\confYear{2025}

\title{\net{}: An Online Motion Model for 4D Heart Reconstruction}

\author{
Yihong Chen\textsuperscript{1} \quad Jiancheng Yang\textsuperscript{1} \quad Deniz Sayin Mercadier\textsuperscript{1} \quad Hieu Le\textsuperscript{1} \quad Pascal Fua\textsuperscript{1}\\
\textsuperscript{1}CVLab, EPFL\\
{\tt\small \{name.surname\}@\{epfl.ch\}}
}

\begin{document}
\maketitle

\begin{abstract}

We present a novel approach to reconstruction of 3D cardiac motion from sparse intraoperative data. While existing methods can accurately reconstruct 3D organ geometries from full 3D volumetric imaging, they cannot be used during surgical interventions where usually limited observed data, such as a few 2D frames or 1D signals, is available in real-time. We propose a versatile framework for reconstructing 3D motion from such partial data. It discretizes the 3D space into a deformable tetrahedral grid with signed distance values, providing implicit unlimited resolution while maintaining explicit control over motion dynamics. Given an initial 3D model reconstructed from pre-operative full volumetric data, our system, equipped with an universal observation encoder, can reconstruct coherent 3D cardiac motion from full 3D volumes, a few 2D MRI slices or even 1D signals. Extensive experiments on cardiac intervention scenarios demonstrate our ability to generate plausible and anatomically consistent 3D motion reconstructions from various sparse real-time observations, highlighting its potential for multimodal cardiac imaging. Our code and model will be made available at \url{https://github.com/Scalsol/MedTet}.

\end{abstract}
    

\section{Introduction} 
\label{sec:intro}

Cardiac 3D modeling, along with motion recovery and simulation, is central to numerous medical applications, ranging from personalized level~\cite{meyer2020genetic} to population scale~\cite{bai2020population}.  Significant progress has been made in anatomical shape modeling~\cite{bai2015bi,ye2023neural,dou2024generative,yang2024generating} and motion generation~\cite{Meng22b,Meng23b,qiao2023cheart,yuan2023myo4d,qiao2024personalised}. Consequently, accurate reconstruction of both static and deforming 3D organ geometries from medical imaging is now well understood~\cite{wang2020deep,Cicek16,Wickramasinghe20,kong2021meshdeform,bongratz2022vox2cortex,cruz2021deepcsr,Yang22a}.


\begin{figure}
    \centering
    \includegraphics[width=\linewidth]{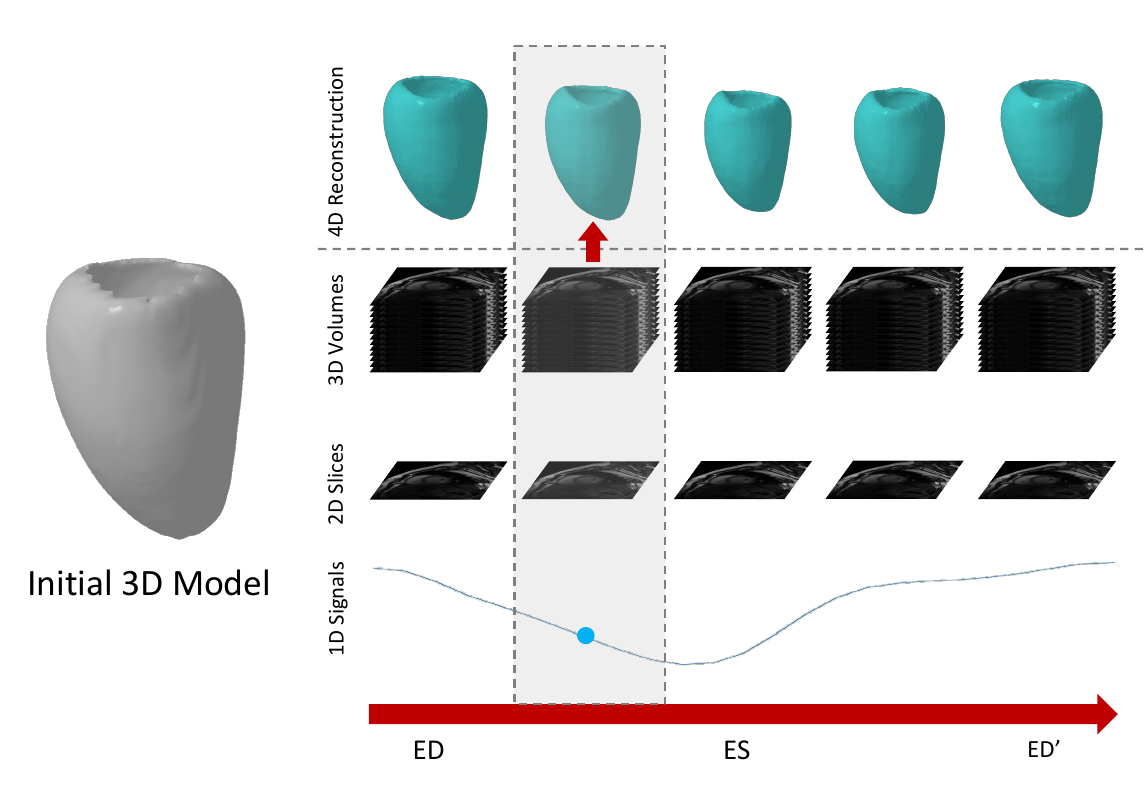}
    \vspace{-6mm}
    \caption{\textit{Online cardiac motion recovery from full or partial observations.} Given an initial 3D model represented using deep tetrahedra~\cite{Shen21a}, our method recovers its shape at time $t$ using either full 3D volumes, a few 2D slices, or even sparser 1D signals, such as time-volume curves.}
    \label{fig:teaser}
    \vspace{-6mm}
\end{figure}

However, a key limitation of most existing methods is their reliance on full volumetric data, which cannot currently be acquired in real-time during an intervention. This limitation precludes their use during surgery, which is the key issue addressed in this paper. In real-world scenarios, the intraoperative data that can be acquired in real time is sparse, such as 2D frames from localized MRI and echocardiography, and 1D signals from electrocardiogram (ECG). This creates a need for a model able to recover full 3D motion from such limited information, so that, unlike earlier methods, it can be used in the operating room. 

To the best of our knowledge, this has never been achieved before, and it presents significant challenges for complex, non-rigid structures undergoing multi-directional motion, especially given the fact that there is little to none annotated data that can be used for training purposes. Therefore, we need a method able to treat sparse 2D slices and 1D signals, as partial observations of a larger 3D space and to use them to guide the deformation of a 3D object and produce 3D motion that aligns with the observation. Additionally, it must ensure that the 4D reconstruction behaves coherently and realistically, maintaining natural motion consistency across the whole structure.

To address this challenge, we propose using a hybrid representation that relies on Deep Marching Tetrahedra (DMTet)~\cite{Shen21a}. As in DMTet, we discretize the 3D space into a deformable tetrahedral grid, in which each vertex is assigned a signed distance value, which is used to estimate a target sign distance function (SDF) in the whole volume. One strength of this approach is that its theoretical resolution is unlimited, unlike that of a voxel-based one,  while being able to model interaction inside the volume enclosed by the object surface, unlike a simple triangulation. Another is that the  implicit component effectively infers and reconstructs missing structures, while the explicit mesh provides control over  motion dynamics, making it possible to ensure that the resulting motion remains both plausible and coherent. In practice, we use it to build a 3D model of the target organ {\it before}  the intervention when there is time to acquire volumetric data and we then train the model in a weakly-supervised manner to enable deformation inference from only the sparse intraoperative data that can be obtained during the intervention. The training being potentially weakly-supervised is essential because annotated full 3D model would be simply not available or costly to obtain.

We conduct extensive experiments on an online heart tracking task, where only a few 2D slices can be acquired at a sufficiently high frame rate. We demonstrate that by incorporating a universal observation encoder and a specialized training pipeline, our model can accommodate varying numbers of 2D slices during inference, with full 3D being a special case--a capability that had not been demonstrated before. Furthermore, we show that our 3D model can also be driven by 1D signals, such as volume size that can also be obtained from other sensors (\eg{}, echocardiography). In other words (\cref{fig:teaser}), our model can condition not only on full 3D volumes when available, but also on sparser 2D slices or even 1D signals, enabling potentials of multimodal capabilities within cardiac imaging and interventions.


\section{Related Works} 
\label{sec:related}

\subsection{Shape and Motion Reconstruction}

3D shape representation, a fundamental aspect of computer graphics and vision, can be broadly categorized into explicit and implicit methods. Explicit methods, including voxel~\cite{Wu15b}, point~\cite{Qi17a,Qi17b,qi2018frustum}, and mesh-based techniques~\cite{Hanocka19,Remelli20b}, directly represent 3D shapes. Voxel-based models align well with volumetric data but are memory-intensive. Point-based methods are efficient and flexible but lack surface connectivity, complicating reconstruction~\cite{Gross11}. Mesh-based models offer compact and detailed representations but struggle with topological changes. Implicit methods, such as signed distance functions (SDF)~\cite{Park19c} and occupancy fields~\cite{Mescheder19,Chen19c}, provide continuous shape representations and excel at modeling complex surfaces, but lack explicit spatial anchors, making it challenging to enforce topological constraints~\cite{le2023enforcing} and handle motion modeling effectively ~\cite{Pumarola21}.

For motion reconstruction, mesh-based techniques are advantageous due to their explicit vertices, which facilitate \textit{forward-flow} deformation~\cite{niemeyer2019occupancy}. This approach allows direct source-to-target manipulation of mesh vertices, resulting in fluid transformations during training and inference. However, mesh-based methods often struggle with accurate shape representation due to fixed topology and sensitivity to initialization. Implicit methods~\cite{deng2021dif,zheng2021dit,yuan2023myo4d,palafox2021npms,Pumarola21} use indirect target-to-source deformation, providing flexibility in motion modeling, but can suffer from inconsistency due to \textit{backward-flow} deformation~\cite{guo2023forward}. Techniques like diffeomorphic flow~\cite{sun2022topology} offer reversible transformations but do not fully address training difficulties.

In medical imaging, shape reconstruction from volumetric data, such as CT and MRI, has evolved from voxel-based approaches~\cite{wang2020deep,Cicek16,yang2021reinventing} to more efficient geometric representations, including point clouds~\cite{ho2021point,yang2021ribseg,jin2023ribseg,hanocka2020point2mesh,nicolet2021large}, meshes~\cite{Wickramasinghe20,kong2021meshdeform,Wickramasinghe22,bongratz2022vox2cortex}, and implicit methods~\cite{cruz2021deepcsr,Yang22a}. Medical motion reconstruction is critical for capturing dynamics within 3D anatomical structures. Many existing methods~\cite{qin2018joint,puyol2018regional,bello2019deep,yu2020foal,bai2020population,loecher2021using,Meng22b} use voxel-wise representations, which are limited by resolution constraints. Recent mesh-based approaches~\cite{Meng22a,Meng23b,qiao2024personalised,dou2024generative} offer improvements but still face challenges such as limited topological flexibility and surface-only representation, restricting their ability to model internal anatomical interactions.

Our approach leverages deep tetrahedra~\cite{Shen21a} for simultaneous shape and motion modeling, combining explicit vertex manipulation for capturing dynamic motion with the advantages of implicit methods for effective shape reconstruction, without sacrificing representation quality or making strong assumptions about target structures. Our study is the first to introduce deep tetrahedral representation into medical image analysis, specifically for 4D heart modeling.

\subsection{Cardiac Shape Model}

Accurate cardiac shape modeling is essential for understanding patient-specific cardiac function and supporting applications such as disease diagnosis~\cite{meyer2020genetic}, cardiac interventions~\cite{li2024solving}, in-silico clinical trials~\cite{abadi2020virtual}, virtual patients~\cite{dou2024generative}, and population imaging~\cite{bai2020population}.

Cardiac shape modeling has been extensively studied in both 3D structures~\cite{bai2015bi,ye2023neural,dou2024generative,yang2024generating} and 4D motion~\cite{Meng22b,Meng23b,qiao2023cheart,yuan2023myo4d,qiao2024personalised}. Mesh-based methods~\cite{Meng23b,ye2023neural,dou2024generative,luo2024explicit,qiao2024personalised} provide a robust framework for capturing cardiac motion through per-vertex deformation, resulting in smooth transformations. However, these methods often lack topological flexibility and are sensitive to initialization, making it challenging to accurately represent intricate shapes. Implicit methods have also been employed for heart shape representation~\cite{yuan2023myo4d,yang2024generating}, excelling at representing complex geometries due to their continuous nature, but lacking explicit spatial anchors, which limits their ability to model dynamic motion effectively.

Most existing methods assume the availability of complete 3D information, which requires a fully acquired 3D volume. In practical settings, such as cine MRI, slices are sequentially acquired during a breath-hold of 4-6 heartbeats, taking around 10 minutes to capture tens of slices—making this unsuitable for real-time applications. For real-time tasks, such as cardiac interventions, only a few localized 2D cine slices can be acquired at sufficient frame rates due to equipment limitations. Multiple modalities in cardiac applications, including echocardiography (2D videos)~\cite{ouyang2020video}, ECG (1D signals)~\cite{li2024hospital} and others, can help guide interventions; nevertheless, obtaining high-resolution 3D volumes in real time remains challenging.

Some previous works may appear similar to our scenario but are fundamentally different. Methods such as~\cite{Meng22b, Meng23b, ye2023neural, yuan2023myo4d} reconstruct 3D structures from sparse 2D or multi-view MRI but assume the availability of at least one low-resolution 3D volume covering the entire 3D space of interest. Other approaches~\cite{qiao2023cheart, qiao2024personalised} are generative motion models conditioned on patient statistics rather than online cardiac observations, which leads to a lack of spatial correspondence. Consequently, existing methods struggle to reconstruct dynamic 3D structures in real surgical settings where only partial observations are available.

The proposed \net{} is designed for 4D motion modeling from online partial observations (Fig.~\ref{fig:teaser}). Our model is the first to leverage sparse 2D frames or even 1D signals to reconstruct full 3D models, offering unprecedented flexibility to enable new applications when different modalities are available beyond full 3D volumes. As a hybrid representation, our approach combines the detailed shape modeling capabilities of implicit methods with the explicit motion control of mesh-based approaches. Moreover, using tetrahedra as a volumetric representation allows for better capturing of internal structures, which is particularly advantageous when only partial observations are available, as evidenced by experimental results presented in Sec.~\ref{sec:main-results}.


\section{Method}

In this section, we introduce {\it MedTet}, our framework that is able to recover 3D motion from \pf{3D, 2D, and even 1D data}.  

\begin{figure*}[t]
    \centering
    \includegraphics[width=1.0\linewidth]{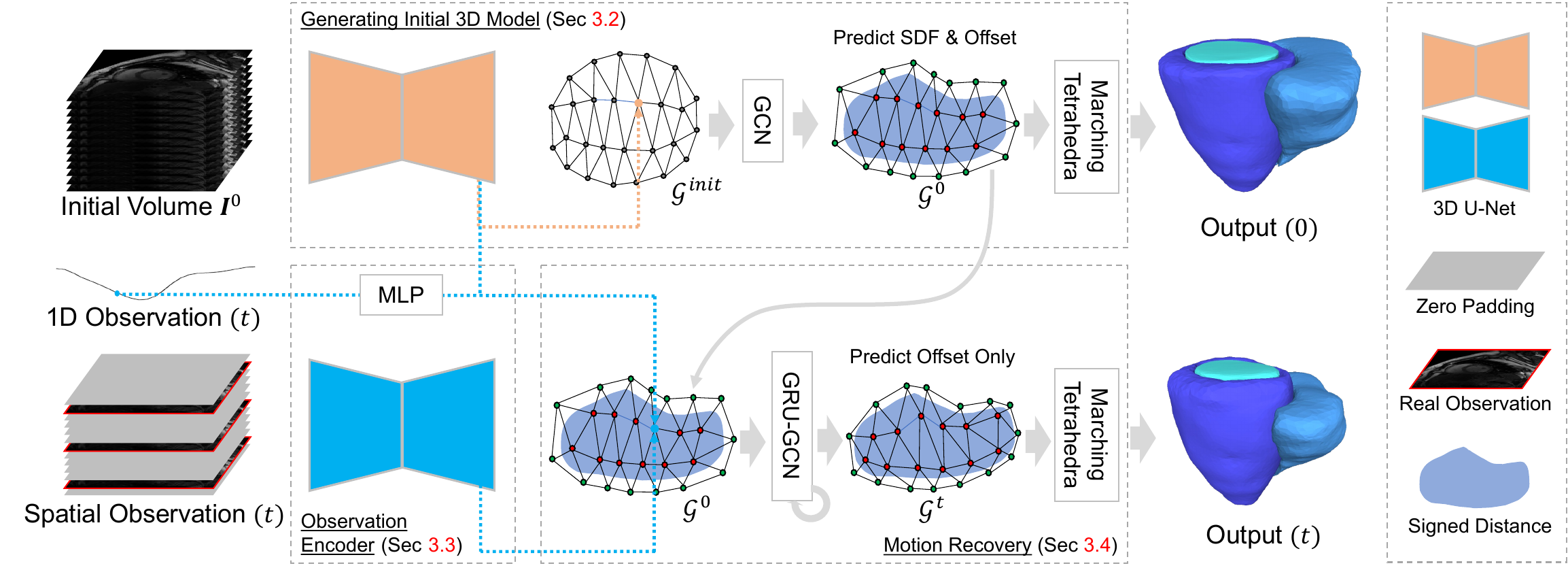}
    \vspace{-6mm}
    \caption{\textit{\net{} recovers cardiac motion from full or partial online observations.} Initially, a 3D image volume is used to generate a tetrahedral 3D model $\mathcal{G}^0$ at $t=0$. Next, when sparse 2D slices are used, unobserved areas are zero-padded to form pseudo-volume, allowing an encoder to handle different number of slices, with full 3D being a special case. When using 1D observations, it is broadcast to each vertex in $\mathcal{G}^0$. Finally, for motion recovery at time $t$, we only predict offsets from $\mathcal{G}^0$, enabling efficient forward-flow deformation.}
    \vspace{-5mm}
    \label{fig:method}
\end{figure*}

We first describe our motion reconstruction pipeline that represents deforming objects as dynamic tetrahedra. Then, we describe our design of the observation encoders to encode information from different signals, with a particular focus on the novel pseudo-3D encoder, which can effectively encode information from a limited number of 2D slices. Finally, we present the training details. The overall framework is given in Fig.~\ref{fig:method}.

\subsection{Formalization}

To parameterize and deform heart models, we use Deep Marching Tetrahedra (DMTet)~\cite{Shen21a}  as a starting point and adapt it for  motion recovery. We discretize the 3D space with a deformable tetrahedral grid, in which each vertex possesses an SDF value. We denote it as $ \mathcal{G} = (\{\mathbf{v}_i\}, \{s_i\}, \mathcal{T})$, where $\mathbf{v}_i$ and $s_i$ are the vertices and their corresponding SDF values in $\mathcal{T}$, the set of all tetrahedrons. The surface could be obtained through marching tetrahedra algorithm~\cite{Akio91}, which is differentiable.


At inference time, in the operating room, we are given an initial model $\mathcal{G}^{0}$ built before the start of the intervention and real time observation $\mathbf{O}^t$ about deformations that have occurred at each subsequent time instant $t$.
We create a new model $\mathcal{G}^{t}$ that accounts for these deformations. It can be done by directly manipulating the vertices of $\mathcal{G}^{0}$. We write
\begin{align}
    \mathcal{G}^{t}&=(\{\mathbf{v}_i^t\}, \{s_i\}, \mathcal{T}) \; , \label{eq:tetMotion} \\
    \mathbf{v}_i^t &=\mathcal{D}(\mathbf{v}_i^0, \mathbf{I}^0, \mathbf{O}^t) \; ,  \nonumber
\end{align}
where $\mathbf{v}_i^0$ is a grid vertex from $\mathcal{G}^0$, $\mathbf{v}_i^t$ is the vertex after deformation, $\mathbf{I}^0$ is the image at first frame, and $\mathcal{D}$ is the deformation model, which we instantiate it in the following sections and makes it possible to recover deformations across the whole sequence.

In our experiments, we simulate the online setting by using volumetric cine MRI sequences $\{\mathbf{I}^t \in \mathbb{R}^{D \times H \times W}\}_{t=0}^T$. We use the first one to reconstruct $\mathcal{G}^{0}$ and we take $\mathbf{O}^t$ to be either 2D slices of $\mathbf{I}^t$ or 1D volume information that can be inferred from it. \pf{In the real scenario, $\mathcal{G}^{0}$ will be acquired from a different sequence but this makes essentially no difference because the same algorithms can be applied to deform it from the intra-operation sequence.}  


\subsection{Generating the Initial 3D Model}
\label{sec:method-static}

We start by encoding the image $\mathbf{I}^0$ and then decoding it into the tetrahedral mesh $\mathcal{G}^{0}$. The image encoding module uses a standard 3D U-Net architecture~\cite{ronneberger15, Isensee21}, with a downsample and an upsample stream. We first create an initial tetrahedral grid $\mathcal{G}^{init}$ by uniformly sampling a unit cube. To decode the U-Net output into accurate vertex coordinates and SDF values at its vertices, we extract feature maps $F^0 \in \mathbb{R}^{C \times D \times H \times W}$ from the last level of the upsample stream, where $C$ is the number of feature channels. We then trilinearly interpolate the feature map at the tetrahedra grids of $\mathcal{G}^{init}$. To create $\mathcal{G}^{0}$, we predict an SDF value and an offset for each vertex using a GCN with the extracted features. 
More formally, let $F^0$ be the feature map encoded from image $\mathbf{I}^0$. The deformed tetrahedral mesh $\mathcal{G}^0$ is expressed in terms of offsets from the regular grid as
\begin{align}
    \mathcal{G}^0 &=(\{\mathbf{v}_i'\}, \{s_i\}, T),   \nonumber\\
     \mathbf{v}_i' &=\mathbf{v}_i+  \Delta \mathbf{v}_i,  \label{eq:tetStatic} \\ 
    (\Delta \mathbf{v}_i, s_i)&=\operatorname{GCN}([F^0(\mathbf{v}_i), \mathbf{v}_i]) \; , \nonumber
\end{align}
where $\mathbf{v}_i$ is a grid vertex of $\mathcal{G}^{init}$ and $\mathbf{v}_i'$ the same grid vertex after deformation. $F^0(\mathbf{v})$ denotes feature extraction at $\mathbf{v}$ from $F^0$ with trilinear interpolation, and $[\cdot, \cdot]$ denotes concatenation: to enhance the network's spatial sensitivity, the vertex location is appended to the feature vector before it is passed to the GCN.

Although reconstructing 3D shapes from static volumes is not our emphasis, our MedTet achieves performance that is comparable to or better than mesh-based and implicit function methods under fair comparison. See Sec.~\ref{sec:analysis} and supplementary materials for details.

\subsection{Observation Encoder}
\label{sec:encoder}

The purpose of the encoder is to turn $\mathbf{O}^t$---3D volumes, 2D slices, or 1D data---into useful features for motion inference. Since our main focus is on using 2D slices, we first describe our design choices for this purpose. We then discuss how they can be adapted to handle the other cases. 

\parag{Spatial Observations: From 2D Slices to 3D Volume.}

Effectively encoding features from $S$ 2D slices $\{\mathbf{I}^t_s \in \mathbb{R}^{H \times W}|s=1,\ldots,S\}$ acquired at timestep $t$ is a challenging task. It would seem natural to use a 2D U-Net to encode information and obtain $S$ 2D feature maps from the 2D slices. However, such an approach has limitations. First, as each feature map is independently encoded by the 2D U-Net, the resulting feature maps lack 3D consistency. Second, treating each slice independently prevents the model from using the available 3D information present in cine MRI sequences during training. 

An alternative is to rely on 3D convolutions, a common choice when seeking to exploit 3D spatial consistency. However, we want to be able to deal with potentially changing numbers of 2D slices at potentially changing locations, which precludes simply stacking the slices. Instead, we create a 3D volume 
$\hat{\mathbf{I}}^t \in \mathbb{R}^{D \times H \times W}$ with the selected slices filled in with actual image data and the others being zero. We then use a 3D U-Net~\cite{Isensee21} to generate a 3D feature map $\hat{F}^t_{3D} \in \mathbb{R}^{C \times D \times H \times W}$, where $C$ is the channel number.
This enables communication between 2D slices while being able to deal with an arbitrary number of them. We further introduce a distillation loss term in \cref{sec:training} to make the feature more discriminative. 

The same pseudo-3D encoder can be used when the complete 3D cine MRI image is available. 

\parag{1D Observations.} For a 1D signal such as the volume of left myocardium $V_{LM}^t \in \mathbb{R}$, we take our inspiration from Transformer~\cite{vaswani2017attention} and  use a 2-layer MLP to encode it as position embedding  $\hat{F}^t_{1D}=PE(V_{LM}^t) \in \mathbb{R}^C$, where $C$ is the feature dimension. For notation simplicity, we assume it is broadcasted to the size of $\mathbb{R}^{C \times D \times H \times W}$.

The reason we choose volume size is that many cardiac function assessments are based on it, such as left ventricular ejection fraction (LVEF). Additionally, the volume-time curve can be obtained from other sensors, like echocardiography~\cite{ouyang2020video}. Our method can be easily extended to incorporate other 1D signals, provided the relevant data availability.

\subsection{Motion Recovery} \label{sec:method-motion}

Given the feature map $F^0$ and tetrahedral mesh $\mathcal{G}^{0}$ obtained from $\mathbf{I}^0$, and feature map $\hat{F}^t$ obtained at time $t$---$\hat{F}^t_{3D}$ when using 2D slices or $\hat{F}^t_{1D}$ when using 1D data----the deformation model $\cal{D}$ of Eq.~\ref{eq:tetMotion} is instantiated as a combination of GCN and GRU layers to effectively aggregate spatial information. We take the update rule to be
\begin{align}
    F^{cat}(\mathbf{v}_i^{(s)})&=[F^0(\mathbf{v}_i^{(s)}), \hat{F}^{t}(\mathbf{v}_i^{(s)})] \; ,\label{eq:feat-extract}\\
    F^{gcn}(\mathbf{v}_i^{(s)})&=\operatorname{GCN}(F^{cat}(\mathbf{v}_i^{(s)}))  \;  ,\label{eq:gcn}\\
    \mathbf{h}_i^{(s+1)}&=\operatorname{GRU}([F^{gcn}(\mathbf{v}_i^{(s)}), \mathbf{v}_i^{(s)}], \mathbf{h}_i^{(s)})  \;  , \label{eq:gru}\\
    \mathbf{v}_i^{(s+1)}&=\mathbf{v}_i^{(s)}+\text{MLP}(\mathbf{v}_i^{(s)}, \mathbf{h}_i^{(s+1)})  \;  . \label{eq:deform}
\end{align}
In Eq.~\ref{eq:feat-extract}, the vertex features are extracted by trilinear interpolation at the corresponding location. The concatenated features are then passed through Eq.~\ref{eq:gcn}-\ref{eq:deform} to gradually update the vertex position, which aggregates the spatial information. Eq.~\ref{eq:feat-extract}-\ref{eq:deform} is repeated $\mathbf{S}$ times, which we set to $2$. 

Notably, \net{} combines the shape modeling strength of implicit methods with the motion modeling advantage of mesh-based approaches. In Sec.~\ref{sec:analysis} and the supplementary materials, we compare mesh, implicit, and tetrahedral methods for joint shape and motion modeling on a synthetic mesh animation dataset. Meshes excel at motion but struggle with detailed shapes, while implicit methods are the opposite. Our hybrid method balances both effectively.



%

\subsection{Training} \label{sec:training}

Creating high-quality annotations for complete sequence is very costly. Thus in many datasets, they are only available for specific keyframes, such as end-diastolic (ED) and end-systolic (ES) phases. This rules out full supervision. Instead, we use the existing annotations to learn to reconstruct in a supervised manner and then learning the deformation model is done in a weakly-supervised way.

\parag{Reconstruction.}
 
 At each training iteration, we randomly choose a labeled image $\mathbf{I}$ with $y^{gt}=\{\mathcal{M}^{gt}, L^{gt}\}$ the ground truth mesh and segmentation annotation. Let $y^p=\{\mathcal{G}^p, L^p\}$ our model's prediction , where $\mathcal{G}^p$ is the output tetrahedra and $L^p$ the segmentation prediction results from the backbone U-Net architecture. We minimize the loss
\begin{align}
    \mathcal{L}_{shape}&(\mathbf{I})=\lambda_{cd}\mathcal{L}_{cd}(\textbf{MT}(\mathcal{G}^p), \mathcal{M}^{gt}) \label{eq:loss-shape} \\
    &+\lambda_{SDF}\mathcal{L}_{SDF}(\mathcal{G}^p, \mathcal{M}^{gt})+\lambda_{ce}\mathcal{L}_{ce}(L^p, L^{gt}) \; , \nonumber
\end{align}
where $\mathcal{L}_{cd}$ is the chamfer distance, $\mathcal{L}_{sdf}$ is an L1-loss used to supervise the predicted SDF values of tetrahedra grid vertices with the ground truth SDF value queried from the ground truth mesh. $\mathcal{L}_{ce}$ is the cross-entropy loss defined on the predicted segmentation map. {\bf MT} demotes the marching tetrahedra algorithm to convert the tetrahedra into surface mesh. The hyperparameters $\lambda_{cd}, \lambda_{sdf}, \lambda_{ce}$ are set to $1.0, 0.1, 0.1$ respectively.


\parag{Motion.}

We then jointly train the observation encoder and the deformation model, while keeping the shape reconstruction model frozen. Take 2D case for example, we first initialize the observation encoder from the shape reconstruction model. At each iteration, we randomly select one sequence $\{\mathbf{I}^t\}_{t=0}^T$, and randomly select a labeled frame $\mathbf{I}^l$ and another frame $\mathbf{I}^u$ from this sequence. After that, we randomly pick 1 to $D$ 2D slices at random location from the labeled frame, and construct the pseudo-3D volume $\hat{\mathbf{I}}^l$. We then generate $F^l_{3D}$ and $\hat{F}^l_{3D}$ from $\mathbf{I}^l$ and $\hat{\mathbf{I}}^l$, respectively. Afterwards, we generate $\mathcal{G}^{u \rightarrow l}$ and $\hat{\mathcal{G}}^{u \rightarrow l}$ using the deformation model Eq. \ref{eq:tetMotion} by taking $F^l_{3D}$ and $\hat{F}^l_{3D}$ as input. This step enables the deformation model to infer accurate motion from the feature map encoded from any number of 2D slices, while also retaining the ability to infer motion from complete 3D feature map. We then train the whole framework by minimizing
%
\begin{align}
    \mathcal{L}_{motion}&(\mathbf{I}^l, \mathbf{I}^u)=\mathcal{L}_{distill}(\hat{F}^{l}_{3D}, F^{l}_{3D}) \label{eq:loss-motion} \\
    &+\mathcal{L}_{cd}(\text{MT}(\mathcal{G}^{u \rightarrow l}), \mathcal{M}^{l}) + \mathcal{L}_{cd}(\text{MT}(\hat{\mathcal{G}}^{u \rightarrow l}), \mathcal{M}^{l}) ,  \nonumber \\
     \mathcal{L}_{distill}&(\hat{F}^l_{3D}, F^l_{3D})=||\hat{F}^l_{3D} - F^l||_2 \;, \nonumber
\end{align}
where $\mathcal{M}^{l}$ is the ground-truth mesh for $\mathbf{I}^l$ and $F^t$ denotes the result by applying U-Net of \cref{sec:encoder} to the full image volume. Minimizing the {\it distillation loss} $\mathcal{L}_{distill}$ ensures that the 3D information from $\mathbf{I}^t$ is successfully distilled into $\hat{F}^t_{3D}$ to make it more discriminative and while minimizing the chamfer losses $\mathcal{G}^{u \rightarrow l}$ and $\hat{\mathcal{G}}^{u \rightarrow l}$ guarantees good reconstruction results for both 2D and 3D conditional input. Thus minimizing $ \mathcal{L}_{motion}$ yields a model that able to infer motion from any number of 2D slices, and also 3D volume.


\section{Experiment}

\begin{table*}[ht]
    \centering
    \begin{tabular}{l|c|ccc|ccc}
        \toprule
         \multirow{2}{*}{Method} & \multirow{2}{*}{Category} & \multicolumn{3}{c|}{4DM (Myo)} & \multicolumn{3}{c}{ACDC (3-Average)} \\
&  & 1-Slice & 3-Slice & Full Volume & 1-Slice & 3-Slice & Full Volume \\
        \midrule
        DeepMesh~\cite{Meng23b} & Explicit (Mesh) & - & - & 0.705 & - & - & 1.402  \\
        MeshDeformNet*~\cite{kong2021meshdeform}& Explicit (Mesh) & 1.123 & 0.909 & 0.702 & 2.776 & 2.245 & 1.388 \\
        DeepCSR*~\cite{cruz2021deepcsr} & Implicit (SDF) & 1.179 & 0.922 & 0.733 & 2.887 & 2.336 & 1.422\\
        \midrule
        Ours w/o P3D encoder & Hybrid (Tetrahedra) & 1.337 & 1.203 & 1.053 & 3.294 & 3.023 & 2.866\\
        Ours w/o distillation loss & Hybrid (Tetrahedra) & 1.155 & 0.920 & \bf 0.695 & 2.933 & 2.413 & 1.362\\
        \hline
        Ours & Hybrid (Tetrahedra) & \bf 1.024 & \bf 0.861 & 0.698 & \bf 2.567 & \bf 2.008 & \bf 1.350\\
        \bottomrule
    \end{tabular}
    \caption{\textit{Quantitative evaluation on 4DM and ACDC dataset.} On 4DM, annotations are available for the \textbf{full training sequences}, and metrics on a single class (Myo) are reported. On ACDC, annotations are available only at \textbf{the ED and ES phases}, and 3-structure averages (Myo, LV RV) are reported. *: Our modified version to enable these methods for motion prediction (Sec.~\ref{sec:exp-settings}). $CD$ in $\times 10^{-3}$.}
    \label{tab:4dm-main}
    \label{tab:acdc-main}
    \vspace{-3mm}
\end{table*}

\subsection{Experimental Settings}
\label{sec:exp-settings}


\begin{figure*}[th]
    \centering
    \includegraphics[width=0.9\textwidth]{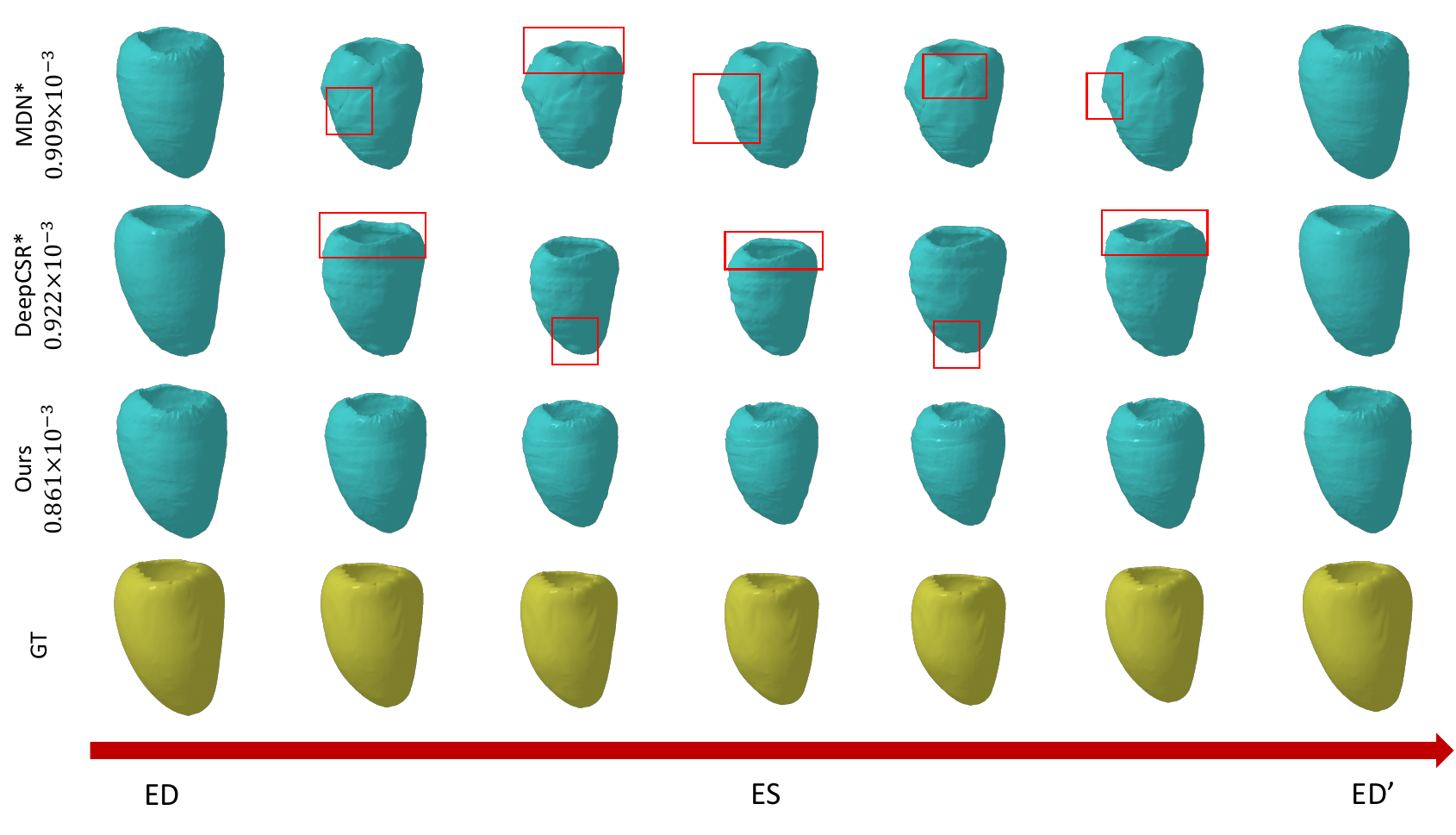}
    \vspace{-3mm}
    \caption{{\it Motion reconstruction results (3-slice) using different methods on 4DM by deforming from ED frame to the target frame.} MeshDeformNet* and DeepCSR* could not faithfully reconstruct the deformation of the object, whereas our approach does.}
    \label{fig:4dm-vis}
    \vspace{-3mm}
\end{figure*}

\paragraph{Datasets and Metrics.} 

We use the 4DM and ACDC datasets~\cite{yuan2023myo4d,bernard2018acdc} for training and evaluation. 
4DM focuses on left myocardium reconstruction. Each subject includes multiple slices (8-10 slices) with a resolution of $1.25 \times 1.25 \times 10$ mm. Each slice covers the video sequence of the cardiac cycle (25 phases). Annotations are available for all frames.
ACDC comprises CMR sequences for  100 patients, each with a slice thickness of 5-10 mm and each sequence covering at least one cardiac cycle, with varying lengths of around 30 frames. It has annotations for 3 classes: Left ventricle/myocardium and right ventricle, but only on the end-diastolic (ED) and end-systolic (ES) frames.
All datasets are randomly split into training, testing, and validation sets in a ratio of 60\%, 20\%, and 20\%, respectively.
More details are provided in the supplementary materials.

For evaluation, we first map pixel coordinates into the range [-1, 1] and use L2-norm chamfer distance (CD) as the evaluation metrics to measure 3D reconstruction quality.

\parag{Implementation Details.}
In all experiments, images are resampled by linear interpolation to a spacing of $1.25 \times 1.25 \times 2$ mm (3D) and $1.25 \times 1.25$ mm (2D). 
For the shape reconstruction model, we use a standard nnU-Net~\cite{Isensee21} with 5 downsample and upsample blocks with channel numbers [32, 64, 128, 256, 320] as the image encoder. We use the same GCN structure as in~\cite{Shen21a} with $3$ layers and $128$ channels. The initial tetrahedra resolution is $128$. For the pseudo-3D encoder, we use the identical network structure as the shape reconstruction model, and the channel number of the embedding MLP is $128$ for 1D experiment. For the deformation model, the hidden dimension of the GCN and GRU are both set to $128$.
We use SGD optimizer with an initial learning rate of 0.01, a weight decay of $3e-5$, and momentum of $0.99$. We train all models for $300/150$ epochs for the reconstruction/motion learning stage. All experiments are conducted on one V100 GPU.

\parag{Baselines.} 

We compare against the following baselines. The * in the names indicate that they are {\it not} the original methods, which were designed to handle static input. The modified versions can accept a small number of slices as input to infer motion. We had to do this because we could not find any existing methods able to do this.

{\bf DeepMesh}~\cite{Meng23b} is a mesh-based method that aims to reconstruct heart motion from full volume sequences. We make two minor modification on it to adapt to our applications. First, we use the ground-truth ED frame mesh as starting point. Second, we use the same cine MRI image sequence as we use as the input to the motion network. As its network structure is specifically designed for motion recovery from full volumes, we only perform the comparison in the full volume setting.
        
{\bf MeshDeformNet*}~\cite{kong2021meshdeform} is a mesh-based method based on improved Voxel2Mesh~\cite{Wickramasinghe20}, originally proposed for static shape reconstruction. We make the necessary adjustments to extend it to the dynamic setting. First,  we replace its original backbone with the same nnU-Net backbone as we use, which improves performance. Second, we use the same pseudo-3D encoder, deformation model and training pipeline as ours for motion prediction. For fair comparison, its mesh template has 11494 vertices for each class, while our output meshes have 6000-14000 vertices, which is in the same range.
   
{\bf DeepCSR*}~\cite{cruz2021deepcsr} is a SDF-based method also designed for static shape reconstruction. Besides using the same backbone, pseudo-3D encoder, deformation model and training pipeline as ours, we incorporate the inference pipeline of~\cite{zheng2021dit} to enable motion modeling with SDF, which has not been explored in the medical imaging field before. For a fair comparison, the grid resolution at test time is set to $128$, the same as that of our initial tetrahedra.

For ablation study purposes, we also compare against the following variants of our own method.

{\bf Ours w/o P3D encoder}. We use 2D U-Net to replace the pseudo-3D encoder as the observation encoder of \cref{sec:encoder}. After obtaining the 2D feature maps, we use an attention mechanism to gather information. Specifically, for each tetrahedra grid, we first extract features at its 2D projection on all feature maps, and then use an attention layer to gather these features.
   
{\bf Ours w/o distillation loss}. We do not use the distillation loss of $\mathcal{L}_{distill}$ of \cref{eq:loss-motion}. 


\begin{figure}
    \centering
    \includegraphics[width=1.0\linewidth]{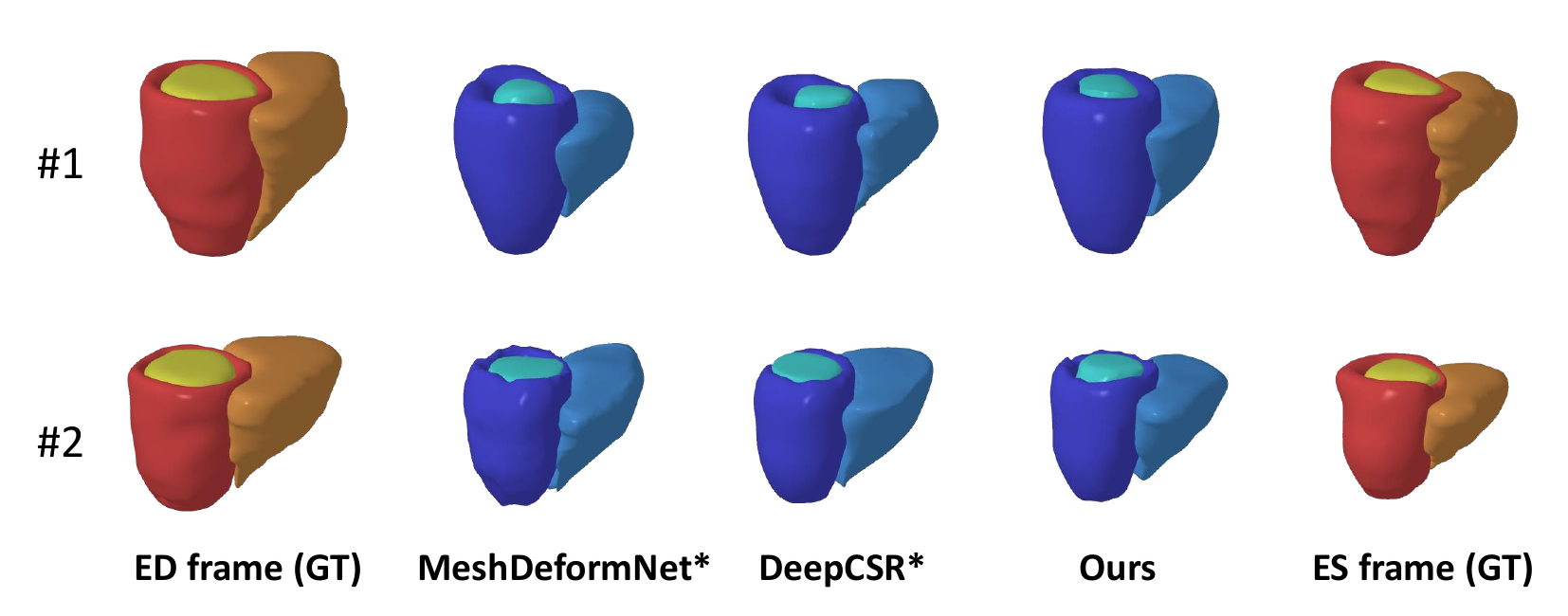}
    \vspace{-3mm}
    \caption{{\it Motion reconstruction results (3-slice) using different methods on ACDC obtained by deforming from ED to ES.}}
    \label{fig:vis-acdc}
    \vspace{-3mm}
    
\end{figure}

\subsection{Comparative Results} 
\label{sec:main-results}

\paragraph{Motion Reconstruction from 2D Slices.} 

We present qualitative results in \cref{fig:4dm-vis} and \cref{fig:vis-acdc} and report out quantitative results  using different numbers of slices in \cref{tab:4dm-main} for the 4DM and ACDC datasets. The 1- and 3- slice columns refer to the number of slices being used to recover the motion. Here, we always use the central 1/3 slices for comparison purpose. In practice, slices at any position could be used. The {\it full volume} column denotes performing the same computation but using all the slices. 

All methods perform similarly well when the full image volume is available. This shows that our framework is also a strong contender for reconstruction accuracy from full volumes, even though that was not our primary focus. When the number of 2D slices is limited, MedTet really comes into its own and significantly outperforms the baselines.  This confirms that the tetrahedra representation used in our model is to better preserve spatial information within the object, which makes it a better choice to infer the 3D motion when only limited information is available.

\parag{Motion Reconstruction from 1D Volume Information.} 

\cref{fig:1d-vis} provides the results of 1D reconstruction experiment on the 4DM dataset. Surprisingly, without any image features, we could still achieve a CD of $1.325\times 10^{-3}$, which is very close to the 1 slice experiment in Table \ref{tab:4dm-main}. From the figure, we can also find the volume fluctuation curve is very close to the ground-truth one. Its problem is that it is not reconstructed at the correct position, which is understandable, as it does not have any location information as input. This demonstrates our method is a strong framework for motion reconstruction from 3D/2D/1D data. 


\begin{figure}
    \centering
    \includegraphics[width=1.0\linewidth]{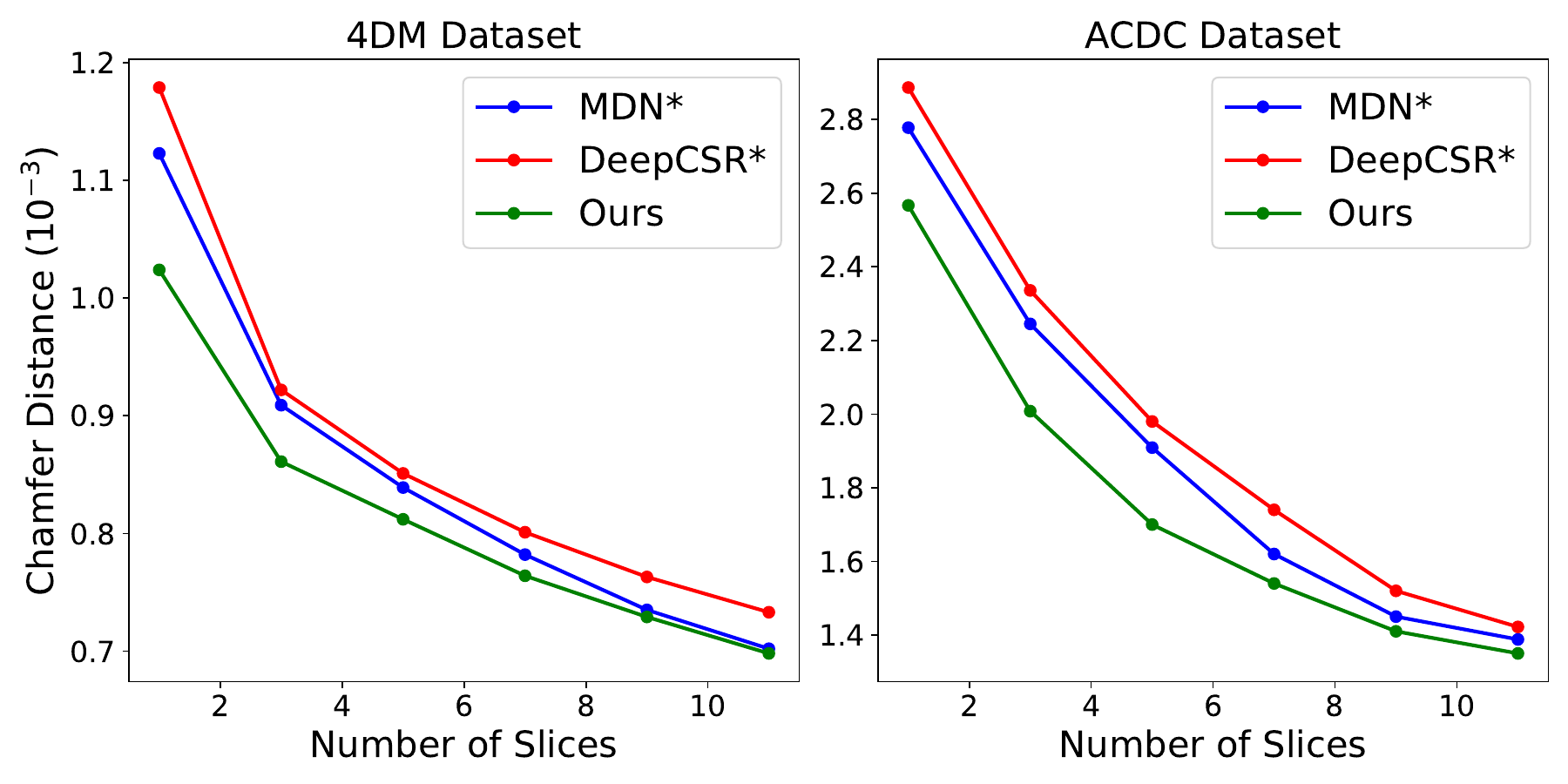}
    \vspace{-5mm}
       \caption{{\it Accuracy as a function of the number of slices.}}
    \label{fig:line-plot}
    \vspace{-3mm}
    
\end{figure}


\begin{figure}
    \centering
    \includegraphics[width=1.0\linewidth]{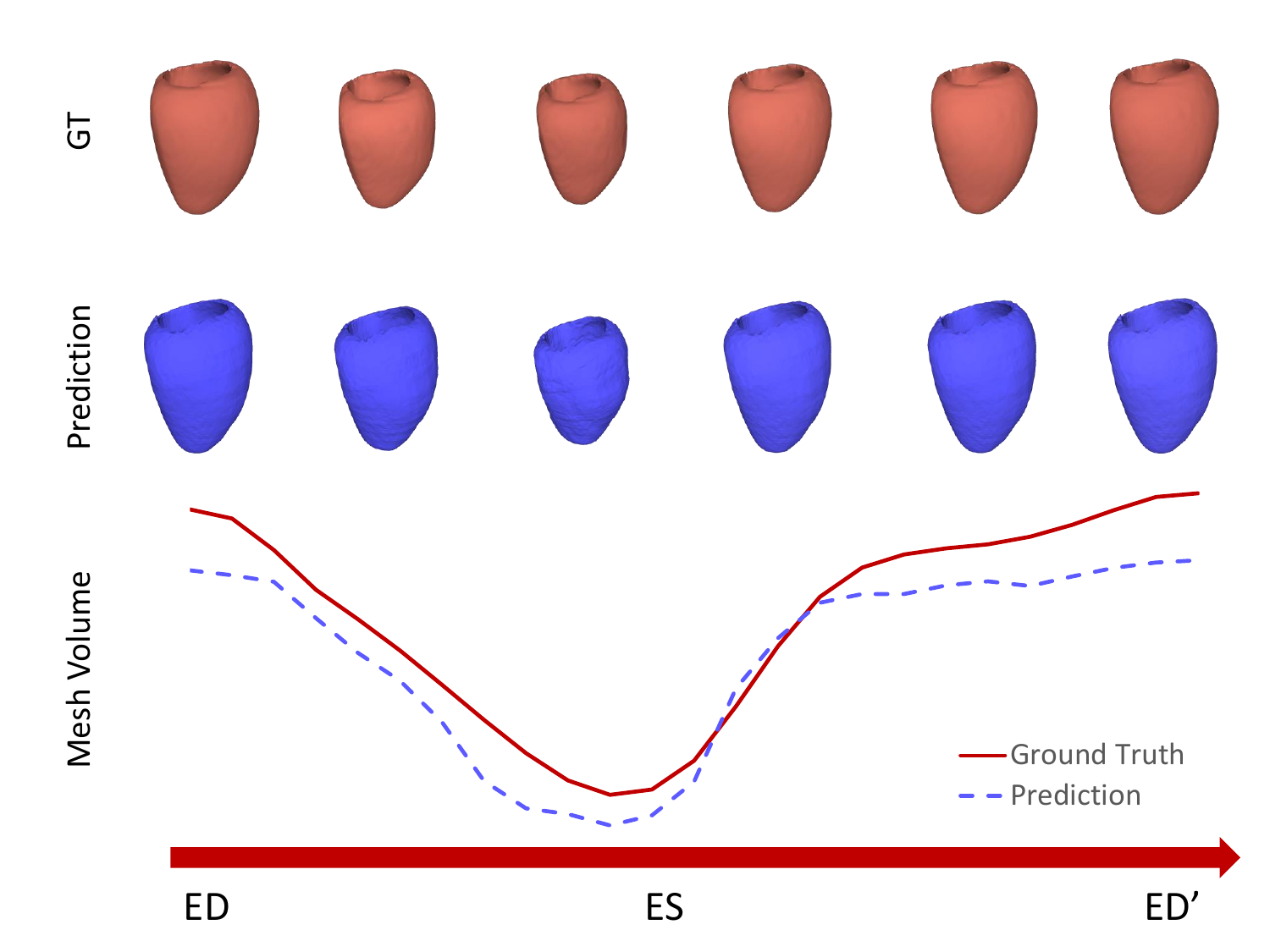}
    \vspace{-6mm}
    \caption{{\it Results of motion reconstruction using only scalar volume values as the conditional input.}}
    \label{fig:1d-vis}
    \vspace{-3mm}
    
\end{figure}

\subsection{Ablation Study and Analysis} 
\label{sec:analysis}

\paragraph{Contribution of Different Components.} 

At the bottom of \cref{tab:4dm-main}, we show what happens when we turn off some components of our approach. The pseudo-3D encoder and the distillation loss are both important for our model's superior performance. The performance drops a lot by replacing the pseudo-3D encoder with 2D encoder, and it can not recover motion even when full volume is given as input. This verifies that only using 2D convolution is not able to provide 3D spatial information to the feature map. Though it may be possible to lift the performance by designing a sophisticated feature gathering process from 2D feature maps, our pseudo-3D encoder is still a simple yet effective choice.

Without the distillation loss, we could not use all the available 3D information from $\mathbf{I}^t$. Thus the feature map is not very discriminative when there are few input 2D slices, resulting in inferior results. When the full 3D volume is available as input, it still produces high-quality results, as earlier methods~\cite{Meng22b, Meng23b, ye2023neural, yuan2023myo4d}.

\parag{Error Analysis.} 

\begin{figure}
    \centering
    \includegraphics[width=1.0\linewidth]{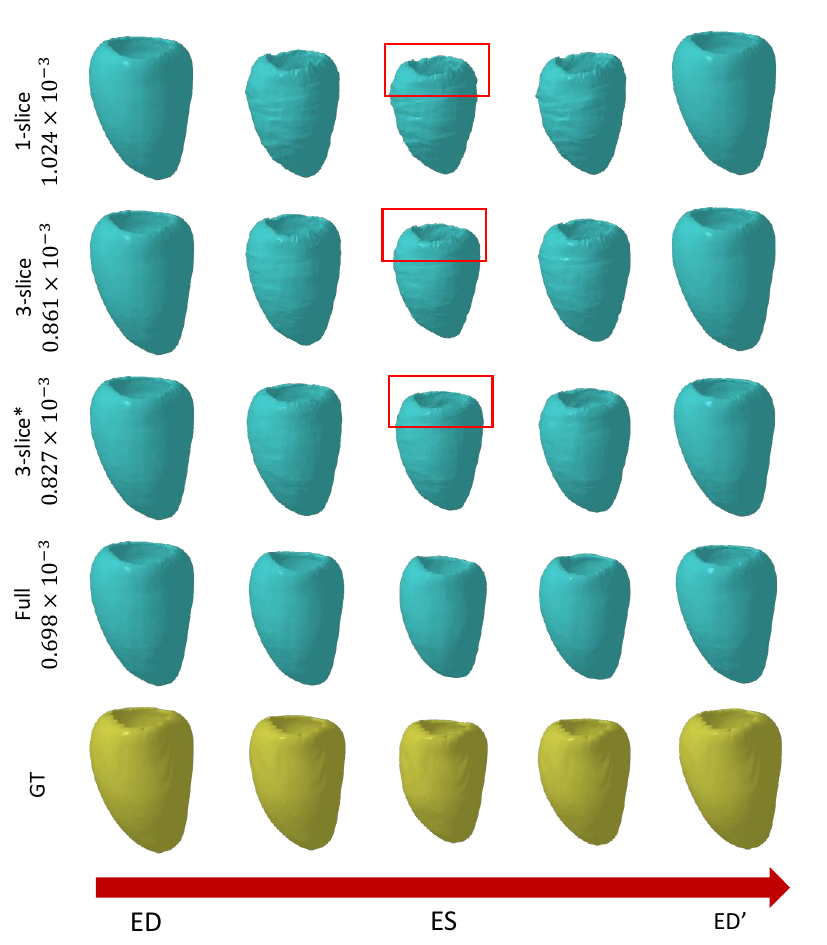}
    \vspace{-6mm}
    \caption{{\it Ablation on the number of slices}. We choose the central 1/3 slices for the first two rows. For the 3rd row, we choose 3 slices from the central 5 slices with stride 2. Chamfer distance is given as reference. Better zoom in for view the details.}
    \label{fig:diff-slice}
    \vspace{-6mm}
    
\end{figure}

In \cref{fig:diff-slice}, we show results when using different numbers of slices. Even when using only the central slice, MedTet captures the heart's general deformation. The most erroneous part is at the top, which is reasonable as it is the most distant part to the position of the input 2D slice. As shown in \cref{fig:line-plot}, the accuracy increases with the number of slices both for our method and for the baselines and MedTet performs best in relation to the others for low numbers of slices. As our model is able to accept 2D slices of any number, at any position as input, it is also interesting to investigate how the choice of slices would influence the result. As shown, if the 3 slices are chosen with a larger stride, we can have better reconstruction results.


\parag{Runtime.} 
As we are targeting real-time interventions, the model runtime is an important consideration. As the inference results of the shape reconstruction could be cached, we only evaluate the runtime of the observation encoder and deformation model. On V100, our model could run at 20 FPS, which is close to the 25 FPS of most current clinical equipment. Also, the inference speed will not be influenced by the number of input 2D slices, as we use the same pseudo-3D encoder. We could further accelerate our model by choosing a lighter encoder or use quantization techniques. We leave this as future work.

\parag{More Visualization Results.} We provide additional visualization and video results in the supplementary materials to help better understand the behavior of our model.


\parag{Static Shape Reconstruction.}
To evaluate our model performance on the static shape reconstruction task, we perform experiments on segmentation datasets CHAOS~\cite{Kavur19}, and MMWHS~\cite{zhuang2019mmwhs}. Traditional approaches in this area usually use triangular mesh or implicit field as their underlying representation. To the best of our knowledge, we are the first to introduce the hybrid tetrahedra representation into the medical imaging domain. We find our method could achieve comparable of even better performance than current SOTA methods. See supplementary materials for details.

\parag{Shape Sequence Learning.} To analyze the ability of joint shape and motion modeling, we conduct experiments that directly reconstruct natural shape sequences on DeformingThings4D~\cite{li20214dcomplete}, a dataset of synthetic mesh animation sequences. We observe that mesh are better for motion modeling but falls short on shape representation, and vice-versa for implicit methods. Our hybrid method achieves a balance on both side and achieve good performance in both tasks. See supplementary materials for details.


\section{Conclusion}

We have proposed a unified framework for reconstructing heart motion in 3D from the limited amount of data---one or more 2D slices of MRI data, or an even weaker signal such as a time-volume curve--available during an intervention. We have demonstrated good performance and shown that it can operate in nearly real time. Furthermore, our weakly supervised approach to training makes it practical in a hospital setting. Finally, there is nothing in our method that is truly specific to the heart. It could be expanded to any other deforming organ, or even beyond medical imaging. 

While achieving promising results, our method still has limitations. First, there remains a performance gap between motion reconstruction using only a few slices versus the full volume. We plan to introduce physical constraints as prior knowledge into the model to reduce this gap. Second, our current framework trains for 2D and 1D signals separately. We intend to investigate their joint use. A natural next step would then be to incorporate more signals, such as ECG, into our framework. We are now working on deploying {\it MedTet} in a real operating room setting in the context of a collaborative project with a teaching hospital.

{
    \small
    \bibliographystyle{ieeenat_fullname}
    \bibliography{bib/string,bib/vision,bib/graphics,bib/biomed,main}
}

\clearpage
\setcounter{page}{1}
\maketitlesupplementary

\section{More Training Details}

\paragraph{Training Pipeline.} 
In the main text, we use 2 datasets: 4DM~\cite{yuan2023myo4d} and ACDC~\cite{bernard2018acdc} for training and evaluation. 4DM dataset has ground truth segmentation and mesh annotation on all frames, so to apply our weakly-supervised training pipeline, we just need to randomly choose 2 frames from the sequence. ACDC dataset only has segmentation annotation at ED and ES phase, so we first obtain ground truth meshes by applying marching cube algorithm to the segmentation annotation. Then at each iteration, we choose either the ED frame or ES frame as $\mathbf{I}^l$, and another random frame as $\mathbf{I}^u$.

\parag{1D Experiment.} We perform this experiment on the 4DM dataset as it provides annotation for all frames. We directly compute the volume of the left myocardium from the ground truth. It is then divided by the volume of cube $[0, 1]^3$ to be normalized to the range of $[0, 1]$. This normalized volume information is then taken as input to the embedding layer which is implemented as a 2-layer MLP. It is a somewhat simplified setting. We conduct this experiment to prove that it is feasible for our framework to infer motion even when sparse 1D information is given as input. In practice, the volume-time curve can be obtained from other sensors, like echocardiography~\cite{ouyang2020video}. We also plan to combine it with 2D signals, or try other 1D signals in future experiments.

\section{More Visualization Results}

\begin{figure*}
    \centering
    \includegraphics[width=1.0\linewidth]{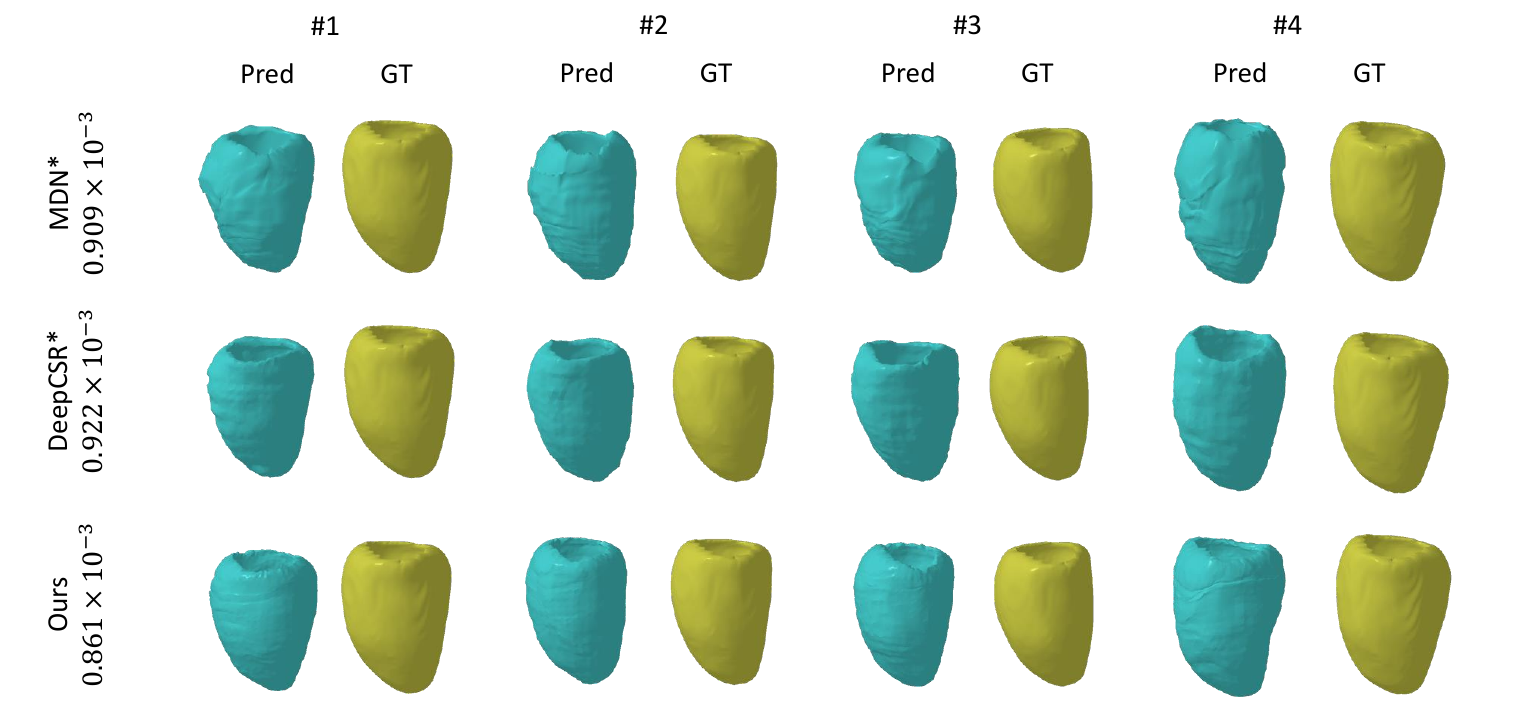}
    \caption{\textit{Pairwise motion reconstruction results (3-slice) comparison using different methods on 4DM by deforming from ED frame to ES frame}. Here we show the results from 4 different sequences. This figure serves as a supplement to Figure 3 of the main text.}
    \label{fig:supp_pairwise_method}
\end{figure*}

\begin{figure*}
    \centering
    \includegraphics[width=1.0\linewidth]{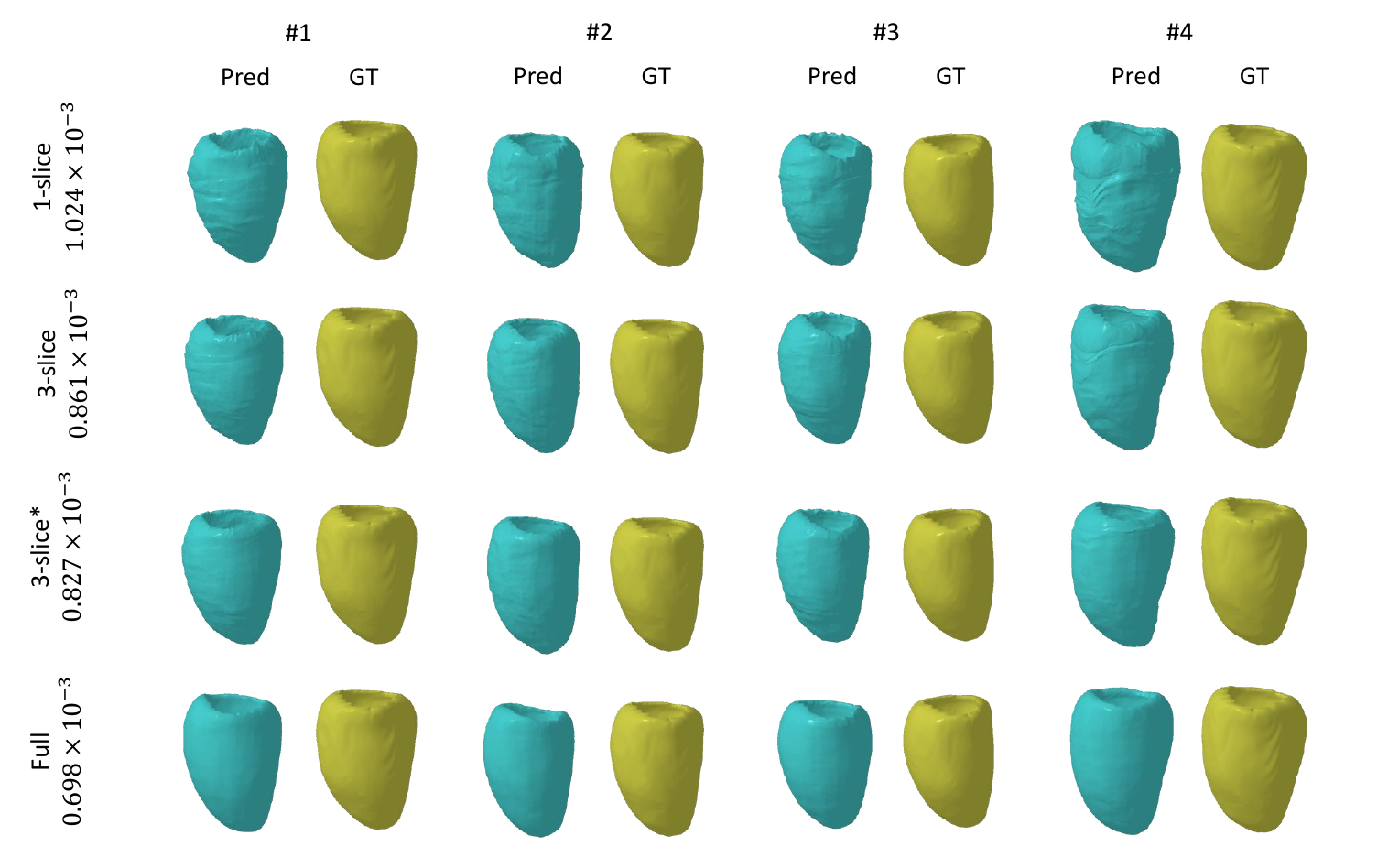}
    \caption{\textit{Pairwise motion reconstruction results comparison on ES frame with different number of slices as input}. Here we show the results from 4 different sequences. This figure serves as a supplement to Figure 7 of the main text.}
    \label{fig:supp_pairwise_slice}
\end{figure*}

\paragraph{Pairwise Comparison Visualization.} In \cref{fig:supp_pairwise_method}, we provide a pairwise motion reconstruction result comparison on ES frame using different method. Here we visualize the prediction results from 4 different image sequences. This figure serves as a supplement to Figure 3 of the main text to help readers better compare the prediction results with ground truth. As can be seen from the figure, MeshDeformNet* and DeepCSR* could not faithfully reconstruct the deformation of the object, whereas our approach could reconstruct the deformation much better. \cref{fig:supp_pairwise_slice} provides the pairwise motion reconstruction result comparison on ES frame using different numbers of slices as input. Here we also visualize the prediction results from 4 different image sequences. This figure serves as a supplement to Figure 7 of the main text. It is evident from the figure that 1) The most erroneous part is at the top and 2) the reconstruction results improve when the number of input slices increases.

\paragraph{Sequence Visualization.} We also provide GIFs results in the project page for whole sequence visualization, which includes: 1) Visualization results of different methods on 4DM 2) Visualization results of different number of slices on 4DM and 3) Visualization results on ACDC. (As ACDC dataset only has annotation on ED and ES frame, the sequence visualization results are presented for demonstration purposes.)

\section{Experiments on Static Shape Reconstruction}

In this section, we provide the experiment results of our tetrahedra-based framework on the static shape reconstruction task. We show that it can achieve performance comparable to or even better than previous methods.

\parag{Task Formulation.} The goal of static shape reconstruction is to reconstruct the shape of interest from a given volumetric image $\mathbf{I} \in \mathbb{R}^{D \times H \times W}$.

\parag{Datasets and Metrics.} CHAOS dataset~\cite{kavur2021chaos} is a dataset consisting of 20 labeled CT images. The task is to segment the liver from the image, which is a single object reconstruction problem. To further demonstrate our model's capacity to handle complex scenarios, we evaluate our model on the MMWHS-CT dataset~\cite{zhuang2019mmwhs} consists of 20 full heart labeled CT images, which contains 7 classes (left ventricle, right ventricle, myocardium, etc.) of objects that need to be classified, which is much harder than the previous one.  We use chamfer distance (CD) as the evaluation metric to measure the reconstruction quality.

\parag{Baselines.} We omit the network structure and training details of ours here as it is already described in Section 3.2, Section 4.1 and Figure 2 (top) in the main text. We compare our method against nnU-Net~\cite{isensee2021nnunet}, MeshDeformNet~\cite{kong2021meshdeform}, Vox2Cortex~\cite{bongratz2022vox2cortex} and DeepCSR~\cite{cruz2021deepcsr}. nnU-Net is a segmentation model that is widely used in the medical imaging domain and is a good representative of SOTA  voxel-based methods. MeshDeformNet and Vox2Cortex are both mesh-based methods. They generate meshes by deforming a pre-defined template. For a fair comparison, we use the same nnU-Net as ours as the backbone. MeshDeformNet and Vox2Cortex use a template with 11494 vertices for each class provided by MshDeformNet while our output meshes have 6000-14000 vertices. We are in the same range.
DeepCSR is an SDF-based method which is trained to regress the SDF value at any given spatial location, and the surface is extracted by the Marching Cube algorithm. For a fair comparison, we also use the same backbone as ours. And the grid resolution at test time is set to 128.

\begin{table}[t]
    \centering
    \begin{tabular}{c|c|c}
        \Xhline{1.0pt}
        Method & CHAOS & MMWHS\\
        \hline
        nnU-Net~\cite{isensee2021nnunet} & 1.644 & 1.991\\
        MeshDeformNet*~\cite{kong2021meshdeform} & 0.6762 & 1.806 \\
        Vox2Cortex*~\cite{bongratz2022vox2cortex} & 0.6584& 1.824 \\
        DeepCSR*~\cite{cruz2021deepcsr} & 0.5698 & \bf 1.748 \\
        \hline
        Ours & \bf 0.5217 & 1.756 \\ 
        \Xhline{1.0pt}
    \end{tabular}
    \caption{\textit{Quantitative evaluation of static shape reconstruction on the CHAOS and MMWHS dataset.} *: We use the same backbone as ours for a fair comparison. $CD$ in $\times 10^{-3}$.}
    \label{tab:static-unified}
\end{table}
\begin{figure*}[th]
    \centering
    \includegraphics[width=0.99\linewidth]{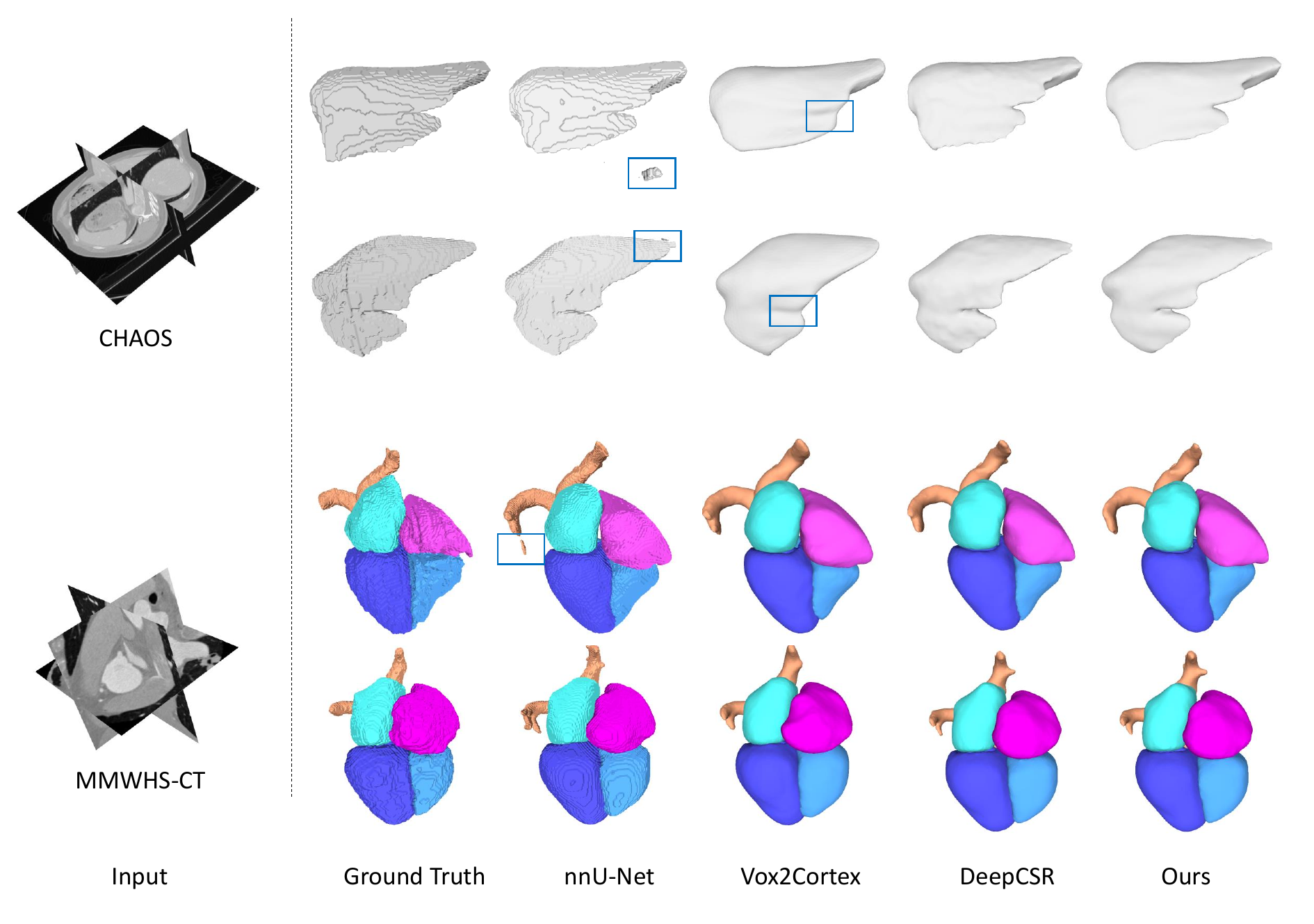}
    \caption{{\it Qualitative results of static shape reconstruction on CHAOS and MMWHS-CT}. nnU-Net could not generate smooth output and has unexpected floaters. Vox2Cortex performs well in general, but fail to capture challenging local details. DeepCSR and our proposed tetrahedra-based framework could obtain a smooth mesh while also being able to capture local details.}
    \label{fig:supp-static}
\end{figure*}

\parag{Results.} As shown in \cref{tab:static-unified}, our tetrahedra-based framework achieves comparable or even better results with other competitors. This validates the effectiveness of our proposed method. \cref{fig:supp-static} is the visualization results. As can be seen from the image, nnU-Net, the voxel-based method could not generate smooth output, moreover, it may generate unexpected floaters due to the lack of surface information. Vox2Cortex could output a smooth mesh as output, but it could not capture good local details due the over-smoothing problem. On the contrary, DeepCSR and ours could obtain a smooth mesh while also able to capture local details.

\section{Experiments on Shape Sequence Learning}

In this section, we compare mesh, implicit, and tetrahedral methods for joint shape and motion modeling on a synthetic mesh animation dataset.

\paragraph{Task Formulation.} Given a 3D mesh sequence:
\begin{equation}
    \{\mathcal{}{O}^{(t)}\}_{t=0,\ldots,T}
\end{equation}
the goal is to jointly optimize a canonical shape model $\mathcal{F}^{(0)}$ and a deformation model $g_\theta^{(t)}$ to represent these objects, i.e., $\mathcal{F}^{(t)}=g_\theta^{(t)}(\mathcal{F}^{(0)})$. Note that here we train and evaluate on the same sequence.

\parag{Our Pipeline.} Similar to Eq. (1) in the main text, suppose $\mathcal{F}^{(0)}=(\{\mathbf{v}_i\}, \{s_i\}, T)$ is the learnt template tetrahedra representation for $O^{(0)}$. We represent $\mathcal{F}^{(t)}$ as:
\begin{equation}
    \mathcal{F}^{(t)}=(\{\mathcal{D}(\mathbf{v}_i, \mathbf{c}^{(t)})\}, \{s_i\}, T) \label{eq:tetmotion}
\end{equation}
where $\mathcal{D}$ is a deformation network and $\mathbf{c}^{(t)}$ is the latent code at timestep $t$. $\mathcal{D}$ can be simply implemented as an MLP layer, but such a simple design could not handle the problem of large deformation. So we implement the deformation model $\mathcal{D}$ in a similar manner as in Section 3.4, which could be written as
\begin{align}
    \mathbf{h}_i^{(s+1)}&=\operatorname{GRU}([\mathbf{c}, \mathbf{v}_i^{(s)}], \mathbf{h}_i^{(s)})  \;  ,\\
    \mathbf{v}_i^{(s+1)}&=\mathbf{v}_i^{(s)}+\text{MLP}(\mathbf{v}_i^{(s)}, \mathbf{h}_i^{(s+1)})  \;  .
\end{align}

We omit timestep $t$ for simplicity. This process is repeated for 3 times.

Suppose $\hat{O}^{(t)}$ is the predicted mesh extracted by applying the differentiable marching tetrahedra algorithm to \cref{eq:tetmotion}, the training loss is defined as 
\begin{equation}
    L=\sum_{t=0}^{t=T}L_{cd}(\hat{O}^{(t)}, O^{(t)}),
\end{equation}
where $L_{cd}$ is the chamfer distance.

\parag{Datasets and Metrics.} We use the DeformingThings4D~\cite{li20214dcomplete} dataset to evaluate the effectiveness of our proposed model. DeformingThings4D is a large-scale synthetic dataset of animation sequences. In our experiment, we select 5 sequences with very different shapes and skeleton structures. We use two types of metrics for comparison. First, we use chamfer distance to measure how the proposed method can match the target shape. Second, as the point correspondence information is known, we additionally measure the point registration performance by comparing the endpoint error (EPE) and strict and relaxed 3D accuracies $Acc_S, Acc_R$ metrics, which is defined as
\begin{align}
    EPE(\{\mathbf{v}_i^s\}, \{\mathbf{v}_i^t\})&=\sum_{i=1}^N\frac{1}{N}||\mathbf{v}_i^{s \rightarrow t}-\mathbf{v}_i^t||_2,\\
    Acc_T(\{\mathbf{v}_i^s\}, \{\mathbf{v}_i^t\})&=\sum_{i=1}^N\frac{1}{N}\mathbbm{1}\left(||\mathbf{v}_i^{s \rightarrow t}-\mathbf{v}_i^t||_2 < T\right)
\end{align}
where $\mathbf{v}_i^s$ and $\mathbf{v}_i^t$ are corresponding vertices of the source and target mesh. $\mathbf{v}_i^{s \rightarrow t}$ is the predicted location of $\mathbf{v}_i^s$ in the target space. $EPE$ measures the average distance between the predicted vertices and the ground truth corresponding vertices, lower is better.
$Acc$ measures the percentage of predicted points that fall within a certain range of the target points, higher is better. $T$ can be $S$ or $R$, represents strict and relaxed 3D accuracies, respectively. In our evaluation, we set $S=0.025$ and $R=0.05$.

\parag{Baselines.} For implicit representation, we use a simplified framework introduced in the Neural Parametric Model (NPM)~\cite{palafox2021npms}. The main focus of this section is to compare the deformation learning ability of implicit representation and our proposed tetrahedra grid-based deformation model. Therefore we leave aside the latent shape model introduced by the approach, and simply compare the capability to learn the non-trivial deformation of a surface. As an oracle, we also conduct an experiment of directly deforming the ground-truth template mesh.

\begin{table*}[t]
    \centering
    \begin{tabular}{l|c|cccc}
        \Xhline{1.0pt}
        Method & $CD_S$ $\downarrow$ & $CD_T$ $\downarrow$ & $EPE$ $\downarrow$ & $Acc_S$ $\uparrow$ & $Acc_R$ $\uparrow$\\\hline
        \color{gray}{Mesh (MLP)} & 
        \multirow{2}{*}{\color{gray}{8.80e-5}} & \color{gray}{1.95e-5} & \color{gray}{0.027} & \color{gray}{72.93} & \color{gray}{83.02} \\
        \color{gray}{Mesh (GRU)}  & & \color{gray}{1.80e-5} & \color{gray}{0.011} & \color{gray}{93.76} & \color{gray}{99.14} \\
        \hline
        Implicit NPM~\cite{palafox2021npms} (MLP) & \multirow{2}{*}{1.85e-5} & 4.76e-5 & 0.040 & 68.34 & 76.86 \\
        Implicit NPM~\cite{palafox2021npms} (GRU) & & 4.34e-5 & 0.028 & 75.13 & 84.50 \\
        \hline
        Ours (MLP) & \multirow{2}{*}{1.98e-5} & 2.15e-5 & 0.036 & 70.77 & 81.06 \\
        Ours (GRU) & & 1.95e-5 & 0.012 & 92.47 & 98.56\\
        \Xhline{1.0pt}
    \end{tabular}
    \caption{\textit{Quantitative evaluation of shape sequence learning on the DeformingThings4D dataset.} MLP: MLP layer as the deformation model, GRU: a GRU layer as the deformation model. $CD_S$: the chamfer distance result of fitting the template source shape. $CD_T$: the chamfer distance result of fitting the target shape. $EPE$: end point error. $Acc_S, Acc_R$: strict and relaxed 3D accuracies.}
    \label{tab:shape-seq}
\end{table*}

\begin{figure}[t]
    \centering
    \includegraphics[width=.99\linewidth]{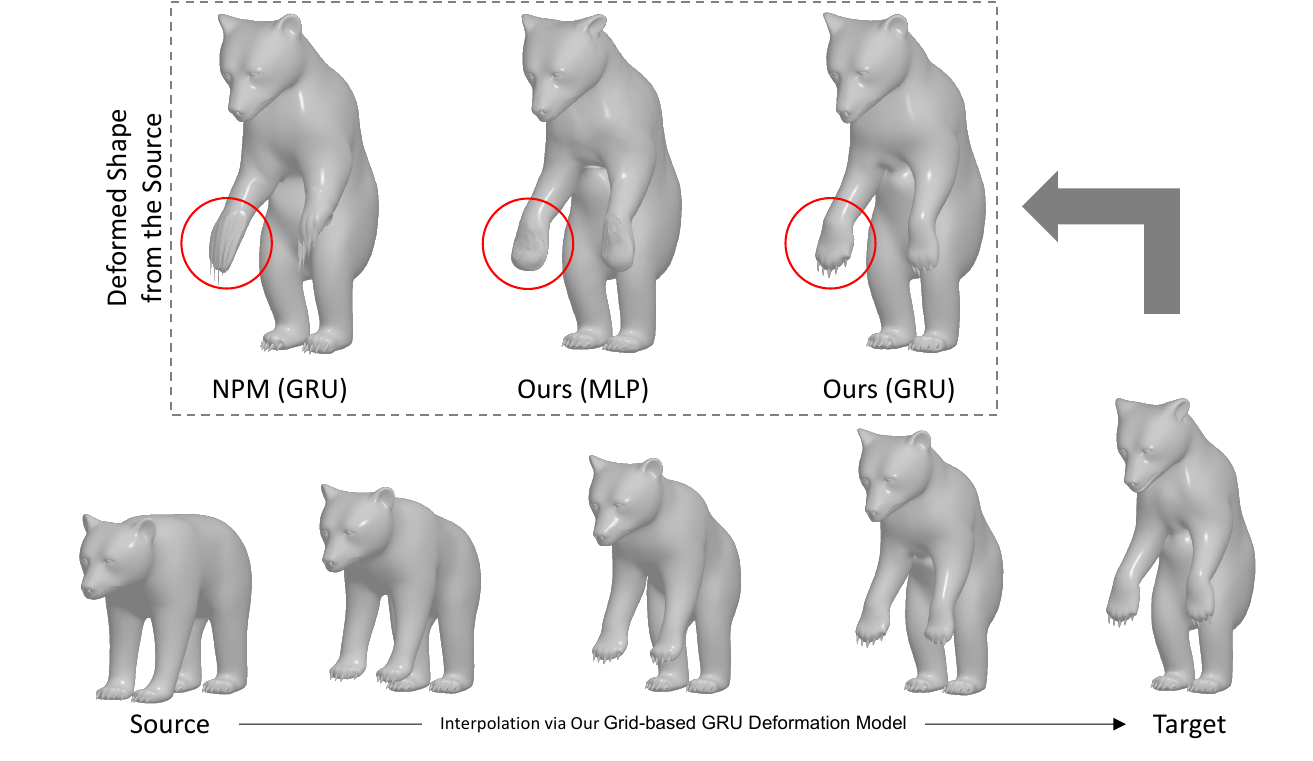}
    \caption{\textit{Qualitative results of shape sequence learning on the DeformingThings4D dataset}. \textit{Top}: Comparison of target shape fitting results of various representation and deformation methods. \textit{Bottom}: Interpolation results.}
    \label{fig:result_shapeseq}
\end{figure}
\begin{figure*}
    \centering
    \includegraphics[width=1.0\linewidth]{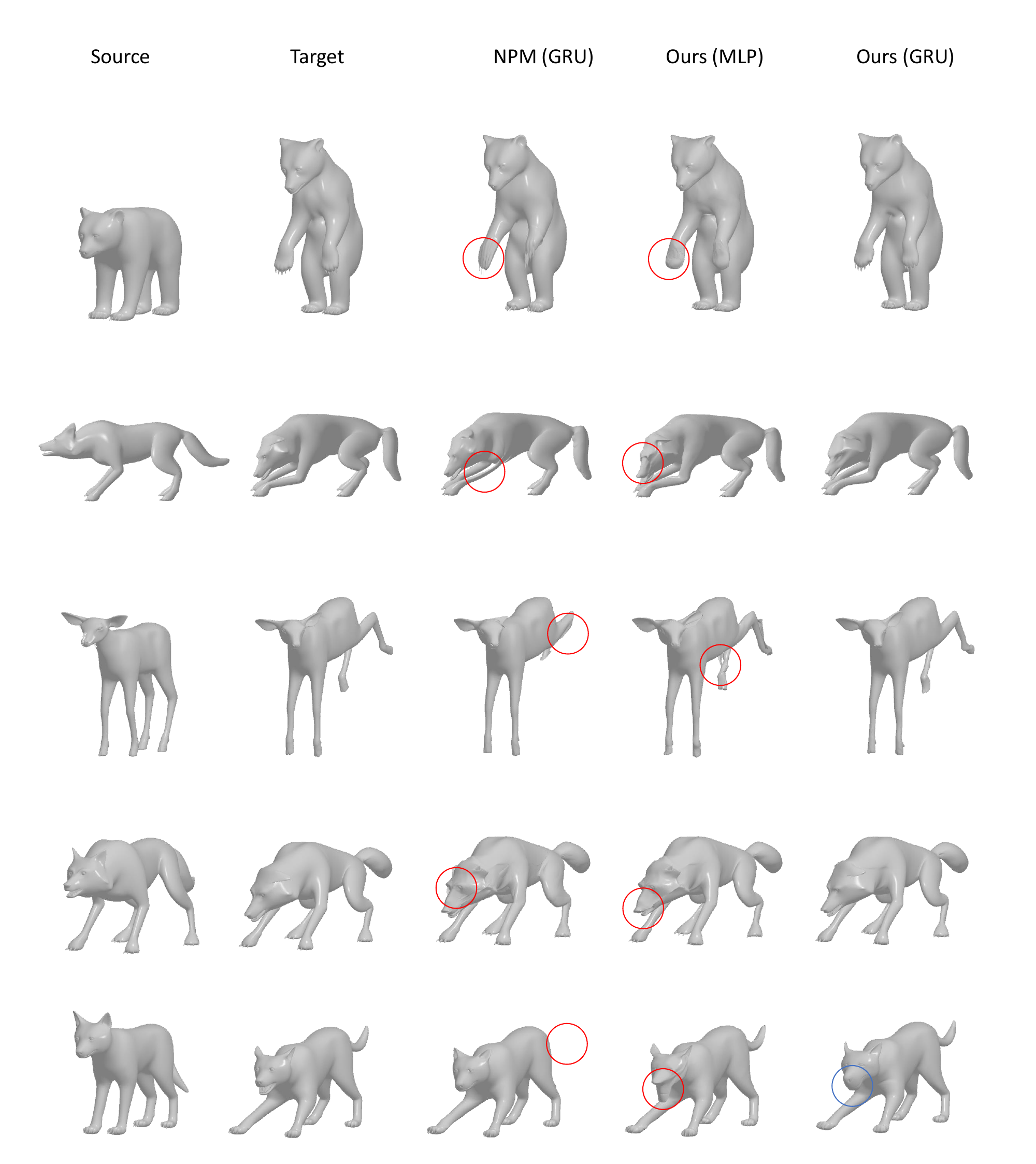}
    \caption{\textit{More qualitative results of shape reconstruction on the DeformingThings4D dataset}.}
    \label{fig:supp_fig1}
\end{figure*}

\parag{Results.} The results are presented in \cref{tab:shape-seq} and the top row of \cref{fig:result_shapeseq}. As can be seen from the table, the implicit baseline fails to learn a plausible deformation field under the scenario of complex motion, resulting in high reconstruction error. With our tetrahedra-based framework, we could achieve a more faithful reconstruction result. 
With a simple MLP layer, though the model could achieve good reconstruction result, it fails to capture correct point correspondence (bear hand), as shown in Figure \ref{fig:result_shapeseq}. The situation gets much improved if we replace the MLP layer with GRU. We find applying GRU to implicit representation is also beneficial, but due to the lack of surface information, it still falls far behind our method.

On the other hand, the oracle study demonstrates that mesh representation is adept at modeling motion. However, as shown in the first column of Table \ref{tab:shape-seq}, mesh is not good at learning to represent complex shapes. Tetrahedra representation performs slightly worse than SDF, while is far better on motion modeling, demonstrating our framework is both good at learning shape representation and motion. More visualization is provided in \cref{fig:supp_fig1}.

This experiment also provides some insights into medical imaging: 1) If the shape to be reconstructed is highly complex, mesh-based methods require a high-quality initial template to perform well, which is not always available. In contrast, our method and implicit methods do not have this limitation. 2) However, implicit methods have shortcomings in motion modeling, making our framework more suitable for image sequence analysis. Furthermore, just as shown in the main text, our framework is more advantageous when observations are incomplete.

\end{document}